\documentclass[10pt,twocolumn,letterpaper]{article}

\usepackage{iccv}
\usepackage{times}
\usepackage{epsfig}
\usepackage{graphicx}
\usepackage{amsmath}
\usepackage{amssymb}

\usepackage{subfig}
\usepackage{booktabs}
\usepackage{multirow}
\usepackage[table]{xcolor}
\definecolor{Graylight}{gray}{0.9}

\usepackage[pagebackref=true,breaklinks=true,letterpaper=true,colorlinks,bookmarks=false]{hyperref}

\iccvfinalcopy 

\def\iccvPaperID{9096} 

\ificcvfinal\fi

\begin{document}

\title{Image Super-Resolution using Efficient Striped Window Transformer}

\author{
Jinpeng Shi$^{1,3}$\textsuperscript{$\dagger$}\textsuperscript{$\ddagger$},
Hui Li$^{1,3}$\textsuperscript{$\dagger$},
Tianle Liu$^{1,2}$,
Yulong Liu$^{1,3}$,
Mingjian Zhang$^{1,3}$, \\
Jinchen Zhu$^{1,3}$,
Dong Liang$^{1}$,
Ling Zheng$^{1}$,
Shizhuang Weng$^{1}$\textsuperscript{$\ddagger$} \\
$^1$Anhui University\\
$^2$University of Science and Technology of China\\
$^3$Fried Rice Lab \\
{\tt\small jinpeeeng.s@gmail.com, weng\_1989@126.com}
}

\maketitle

\ificcvfinal\thispagestyle{empty}\fi

\begin{abstract}
Transformers have achieved remarkable results in single-image super-resolution (SR). However, the challenge of balancing model performance and complexity has hindered their application in lightweight SR (LSR). To tackle this challenge, we propose an efficient striped window transformer (ESWT). We revisit the normalization layer in the transformer and design a concise and efficient transformer structure to build the ESWT. Furthermore, we introduce a striped window mechanism to model long-term dependencies more efficiently. To fully exploit the potential of the ESWT, we propose a novel flexible window training strategy that can improve the performance of the ESWT without additional cost. Extensive experiments show that ESWT outperforms state-of-the-art LSR transformers, and achieves a better trade-off between model performance and complexity. The ESWT requires fewer parameters, incurs faster inference, smaller FLOPs, and less memory consumption, making it a promising solution for LSR. The Code is available at \url{https://github.com/Fried-Rice-Lab/FriedRiceLab}.
\end{abstract}

\begin{figure}[thbp]
    \centering
    \includegraphics[scale=.36]{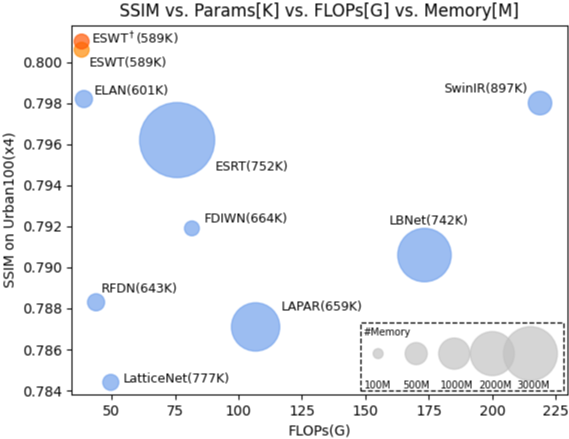}
    \caption{
		Qualitative trade-off comparison between model performance and complexity of $\times 4$ image SR on the Urban100 \cite{Urban100} dataset. The proposed ESWT is marked in orange. The comparison results demonstrate the superiority of our method. ``$\dagger$'' means the use of flexible window training strategy.
    }
    \label{fig: trade_off}
\end{figure}

\let\thefootnote\relax
\footnotetext{
	\textsuperscript{$\dagger$} Co-first authors.
	\textsuperscript{$\ddagger$} Co-corresponding authors.
}

\section{Introduction}
High-resolution (HR) images are essential in various domains, such as satellite imaging \cite{sr1}, medical imaging \cite{sr2}, and surveillance monitoring \cite{sr3}. However, the acquisition of HR images is often constrained by hardware limitations, which leads to low-resolution (LR) images. Single-image super-resolution (SR) is aimed at the reconstruction of HR images from their LR counterpart, which is a challenging problem due to the non-unique mapping between HR and LR images. In recent years, SR transformers, including IPT \cite{IPT}, EDT \cite{EDT}, and HAT \cite{HAT}, have shown great promise in addressing this problem. Nevertheless, the extensive model size and high computational cost of these methods limit their real-world applications.

To achieve a better balance between model performance and complexity, a number of lightweight SR (LSR) transformers have been proposed. SwinIR provides a light-weight version that utilizes window self-attention (WSA) and shifted window mechanism to decrease the computational complexity of transformers. ESRT \cite{ESRT} computes WSA within the overlapped windows to achieve a balance between model parameters and long-term dependent modeling capabilities. ELAN \cite{ELAN} simplifies the computation operation of WSA and introduces an attention sharing mechanism to maintain a performance comparable to that of SwinIR while exhibiting a lower model complexity. LBNet \cite{LBNet} utilizes a recursive mechanism to enhance feature representation capability by increasing the depth of the transformer without adding other parameters. Although LSR transformers have made remarkable progress, existing designs remain  inefficient. Shifted WSA has limited capability to capture long-term dependencies \cite{CCNet, CSwin}, whereas overlapped WSA entails high computational costs \cite{ESRT, HAT}. Recursive mechanisms can reduce model complexity but at the expense of slower inference. Furthermore, serval works have shown that the normalization layer may penalize the generalization capability \cite{b-v1, b-v2} and introduce noise \cite{StyleGAN2, ESRGCNN} to models with limited capacity. LSR transformers are often very shallow, and thus the influence of normalization layers  needs to be investigated.

To further reduce the model complexity and improve the quality of SR images, we propose an efficient striped window transformer (ESWT). We first revisit the normalization layer in the shallow transformer and introduce a concise and efficient transformer structure. Moreover, we propose a striped window mechanism that allows ESWT to model long-term dependencies efficiently. Considering that high-quality SR images originate from the joint optimization of model and training strategy, we also propose a novel flexible window training strategy. This strategy enables ESWT to model long-term dependencies progressively and thus better explore contextual information. The quantitative trade-off comparison of model performance and complexity (Figure \ref{fig: trade_off}) demonstrates the superiority of the proposed ESWT.

The contributions of this paper are summarized as follows:
\begin{itemize}
    \item[$\bullet$] We revisit the normalization layer in the shallow transformer and design a concise and efficient transformer structure to build the proposed ESWT.
    \item[$\bullet$] We propose a striped window mechanism, that can explore contextual information efficiently and has a low computational complexity.
    \item[$\bullet$] We propose a novel flexible window training strategy enables further exploration of the potential of the proposed ESWT without additional cost.
\end{itemize}

\section{Related Works}

\paragraph{Normalization Layer in Transformers}
Normalization layers are common components in transformers. They are typically applied before self-attention (SA) and multilayer perceptron (MLP) to mitigate the effects of internal covariate bias and reduce the dependence of gradients on model size to stabilize training and enhance model generalization. Several works \cite{deeper, ViT, BERT} have demonstrated the importance of normalization layers for training deep transformers with hundreds of layers. However, the transformers used for LSR \cite{ESRT, LBNet, ELAN} are very shallow, typically with only a few tens of layers. The role of normalization layers in such shallow transformers has not been fully explored yet.

\paragraph{Efficient SA}
SA is widely used in SR transformers \cite{SwinIR, HAT, IPT, ELAN, EDT}. However, the computational complexity of SA is quadratic to the image size, which introduces high model complexity for these transformers and affects their further practical applications. For the reduction of model complexity, SwinIR \cite{SwinIR} performs SA in local windows to achieve linear computational complexity. ELAN \cite{ELAN} applies SA by group, which further reduces the model complexity while maintaining a performance comparable to that of SwinIR. However, calculation of SA within the local window causes difficulty in modeling long-term dependencies. To solve this problem, some methods adopt shifted window mechanism \cite{ESRT, LBNet} or overlapped window mechanism \cite{ESRT, HAT}, which either indirectly solves this problem or introduces additional computational cost, respectively.

\paragraph{Training Strategies for LSR Methods}
Numerous training strategies have been proposed to further improve the performance of models on image SR. RCAN-it \cite{RCAN-it} fine-tunes pre-training weights to achieve a model performance that is comparable to that of scratch training, while keeping a lower training cost. EDT \cite{EDT} demonstrates a correlation between different image restoration tasks, such as image SR, deraining, and denoising. Based on this correlation, EDT proposes a related task training strategy to further improve the feature representation of the model. RLFN \cite{RLFN} proposes a three-stage training strategy that accelerates model convergence and improves the quality of SR images. All of these methods demonstrate that a better training strategy can improve the model performance.

\section{Methodology}
In this section, we first revisit the normalization layer in shallow transformers and propose a concise and efficient transformer structure. Subsequently, we propose a striped window mechanism that enables transformers to model long-range dependencies efficiently. Finally, we investigate the feature representations of the ESWT using different striped windows and propose a flexible window training strategy.

\begin{figure*}[thbp]
    \centering
    \includegraphics[scale=.44]{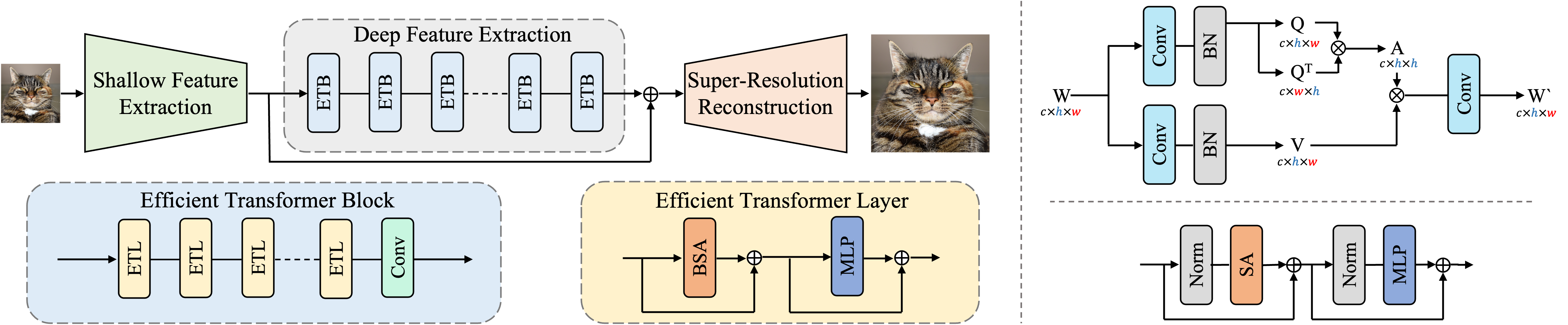}
    \caption{
    	\textbf{Left}: Overall architecture of the proposed ESWT.
    	\textbf{Upper right}: Illustration of SA in a local window $W$ of size $c \times h \times w$ using the BN-embedded SA.
    	\textbf{Lower right}: Illustration of the original TL, in which various normalization layers can be applied.
    }
    \label{fig: all}
\end{figure*}

\subsection{Towards Concise Structure} \label{sec: 3.1}
The transformer layer (TL) is the basic module of trans-formers\cite{SwinIR, SwinT, CSwin, HAT, EDT, ELAN}. As shown in the lower right of Figure \ref{fig: all}, an original TL consists of a SA and a MLP. The normalization layer and residual connection are applied before and after SA and MLP, respectively. To clarify the impact of the normalization layer on shallow transformers, following previous works \cite{ESRT, LBNet, ELAN}, we build a shallow LSR transformer with model size around $600K$ based on the SwinIR structure. To provide a more detailed analysis, we evaluate the performance of the LSR transformer with different normalization layers across various learning rates given that normalization can significantly influence the training strategy of the model (see Section \ref{sec: 4.1} for details).

Although shallow transformers can be trained successfully without normalization layers, our experiments show that normalization layers are crucial for achieving a better balance between model performance and training cost. In addition, considering the limited model capacity, batch normalization (BN) \cite{bn} enables shallow transformers to gain better generalization capabilities by allowing transformers to capture generic features in the mini-batch than layer normalization (LN) \cite{LN}. Combining the above findings and inspired by ELAN \cite{ELAN}, we remove the normalization layer from TL and embed BN into SA to build an efficient and concise transformer structure.

\paragraph{BN-embedded SA}
WSA is widely used in computer vision transformers. WSA partitions an input feature $F \in \mathbb{R}^{C \times H \times W}$ into local windows $W \in \mathbb{R}^{c \times h \times w}$, and uses SA to model dependencies within each local window separately. The upper right of Figure \ref{fig: all} shows the process of modeling dependencies in a local window $W$ using BN-embedded SA (BSA). The BSA initially computes the query matrix $Q$ and the value matrix $V$ using two $1 \times 1$ convolutional layers. BN is applied after each convolutional layer to provide the transformer a more stable training and more powerful generalization capability \cite{ELAN}. This process can be expressed as follows:
	\begin{equation}
		\begin{aligned}
			&Q = H_{BN_{Q}}(K_{Q}W),
			&V = H_{BN_{V}}(K_{V}W),
		\end{aligned}
	\end{equation}
where $K_{Q}$ and $K_{V}$ are the kernel of $1 \times 1$ convolutional layers used for the computation of $Q$ and $V$, respectively; $H_{BN_{Q}}(\cdot)$ and $H_{BN_{V}}(\cdot)$ represent the function of different BNs. In addition, BN can be embedded into its preceding convolutional layer to accelerate the inference of the model further.

Subsequently, the BSA uses matrix $Q$ to compute attention matrix $A$. Key matrix $K$ is not used here to streamline the transformer further. Finally, the BSA uses a $1 \times 1$ convolutional layer to map the matrix $V$ weighted by matrix $A$ to the desired feature space to obtain the output $F_{out} \in \mathbb{R}^{c \times h \times w}$. This process can be expressed as follows:
	\begin{equation}
		F_{out} = H_{Conv}(\frac{H_{Softmax}(QQ^{T})}{scale}V),
	\end{equation}
where $H_{Softmax}(\cdot)$ represents the function of Softmax, $H_{Conv}(\cdot)$ denotes the function of the convolutional layer at the tail of the BSA, and $scale$ is a constant used to control the magnitude of matrix $A$.

\paragraph{Overall Structure}
We build our ESWT based on the efficient and concise structure discussed above. As shown in the left of Figure \ref{fig: all}, the proposed ESWT consists of three main modules: shallow feature extraction module (SFEM), deep feature extraction module (DFEM), and SR reconstruction module (SRRM). For a given LR image $I_{LR}$, the ESWT first converts it from color space to feature space using SFEM to extract the shallow feature $F_{s}$. This process can be formulated as follows:
	\begin{equation}
		F_{s} = H_{SFEM}(I_{LR}),
	\end{equation}
where $H_{SFEM}(\cdot)$ represents the function of SFEM.

Subsequently, the ESWT extracts deep feature $F_{d}$ from $F_{s}$ using DFEM. This module contains $n$ efficient transformer blocks (ETBs) and uses them to extract deeper features block by block. This process can be formulated as follows:
	\begin{equation}
		\begin{aligned}
			F_{d} &= H_{DFEM}(F_{s})\\
		      	&= H_{ETB_{n}}(H_{ETB_{n-1}}(\cdots H_{ETB_{1}}(F_{s}) \cdots)),
		\end{aligned}
	\end{equation}
where $H_{DFEM}(\cdot)$ represents the function of DFEM, and $H_{ETB_{n}}(\cdot)$ refers to the function of $n$-th ETB in DFEM.

\begin{figure}[ht]
	\centering
		\subfloat[Modeling long-term dependencies \textbf{with} striped window mechanism]{
		\includegraphics[width=.45\textwidth]{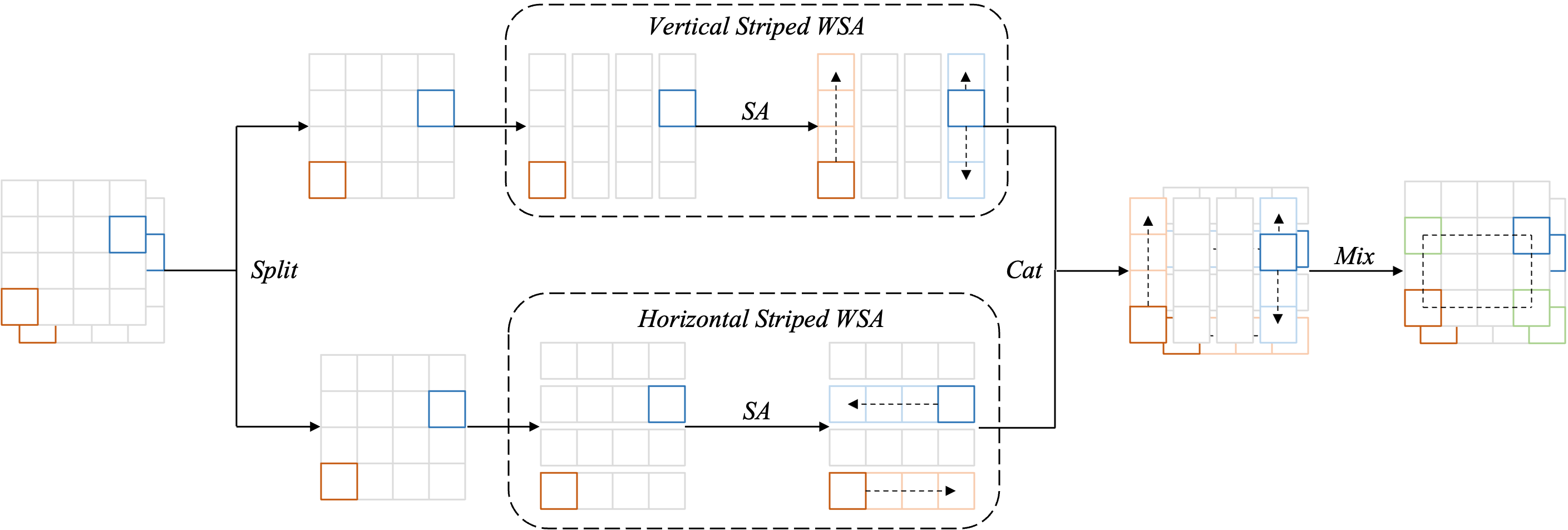}
		\label{fig: strip_wsa}
	}
	\vspace{0.5em}
	\\
	\subfloat[Modeling long-term dependencies \textbf{whithout} striped window mechanism]{
		\includegraphics[width=.45\textwidth]{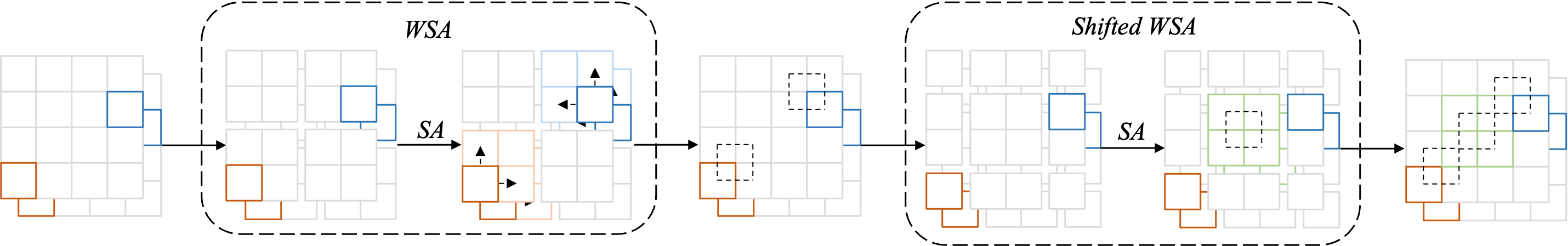}
		\label{fig: shift_wsa}
	}
	\caption{
		Modeling long-term dependencies with/without striped window mechanism. Take the blue and orange positions in the input feature as an example, using the striped window mechanism can model the dependency between the two positions more efficiently. This finding proves the advantage of the striped window mechanism in modeling long-term dependencies. For concise illustration, the size of the input feature is set to $4 \times 4$, and local windows are set to $1 \times 4$ and $2 \times 2$ .
	}
\end{figure}

The ETB contains $m$ efficient TLs (ETLs). In addition, a convolution layer is applied to the tail of each ETB to introduce inductive bias in the transformer \cite{SwinIR}. The function $H_{ETB_{n}}(\cdot)$ can be further expressed as follows:
	\begin{equation}
		\begin{aligned}
			F_{n} &= H_{ETB_{n}}(F_{in})\\
		      	&= H_{C}(H_{ETL_{n, m}}(H_{ETL_{n, m - 1}}(\cdots H_{ETL_{n, 1}}(F_{n - 1}) \cdots))),
		\end{aligned}
	\end{equation}
where $H_{ETL_{n, m}}(\cdot)$ represents the function of $m$-th ETL in $n$-th ETB; $F_{n - 1}$ and $F_{n}$ are the input and output of $n$-th ETB, respectively.

Finally, the ESWT converts $F_{s}$ and $F_{d}$ from feature space to color space using SRRM to reconstruct the SR image $I_{SR}$. By transmitting $F_{s}$, which contains rich low-frequency information, directly to SRRM, the transformer can focus more on reconstructing the lost high-frequency information. The process of SR image reconstruction can be expressed as follows: 
	\begin{equation}
		I_{SR} = H_{SRRM}(F_{s} + F_{d}),
	\end{equation}
where $H_{SRRM}(\cdot)$ represents the function of SRRM.

\subsection{Efficient Long-term Dependency Modeling} \label{sec: 3.2}
Despite WSA makes transformers more cost-effective \cite{SwinT, SwinIR}, it also weakens their capability to model long-term dependencies across local windows. To overcome this issue, some transformers use the shifted or overlapped window mechanism. However, this makes them require stacking more TLs to reach global reception field or need additional computational costs, both of which are detrimental to lightweight models. Inspired by previous work \cite{CCNet, CSwin}, we propose a striped window mechanism to model long-term dependencies efficiently by capturing contextual information from different dimensions in a targeted manner.

\paragraph{Striped Window Mechanism}
The use of striped window mechanism to model long-term dependencies between the blue and orange positions is shown in Figure \ref{fig: strip_wsa}. The input feature $F_{in}$ is split equally into two independent features along the channel dimension. Then, WSAs with $(h, w)$ vertical or $(w, h)$ horizontal striped windows are applied to the two features. This method allows the establishment of in-window dependencies over a larger range of specific dimensions for further exploration of contextual information. Finally, the two features are concatenated along the channel dimension. Given that the vertical and horizontal striped windows are boundary-crossed, a $1 \times 1$ convolutional layer is used to blend the in-window dependencies therein. As a result, the transformer models the long-term dependencies between blue and orange locations more effectively.

\paragraph{Complexity Analysis of Modeling Dependencies}
The computational complexity of modeling dependencies in a local window using the BSA discussed in Section \ref{sec: 3.1} is as follows:
	\begin{equation}
		\Omega = 3c^2hw + 2c(hw)^2,
		\label{equ: 1}
	\end{equation}
where $h$, $w$, and $c$ are the height, width, and channel number of the local window $W$, respectively. According to \cite{SwinT, CSwin}, we omit BN and Softmax here to simplify the analysis.

Based on the above discussion and Equation \ref{equ: 1}, the computational complexity of model long-term dependencies between the blue and orange positions using striped window mechanism is as follows:
	\begin{equation}
		\begin{aligned}
			\Omega &= \frac{HW}{hw} \times [2 \times (2(\frac{C}{2})^2hw + 2(\frac{C}{2})(hw)^2) + C^2hw]\\
		              &= (2C + 2N)CHW,
		\end{aligned}
	\end{equation}
where $N = h \times w$ denotes the total number of pixels in a local window. The length $h$ and width $w$ of the striped window can be adjusted freely to trade-off between model performance and complexity.

Compared with the striped window mechanism, the long-term dependence between the blue and orange positions cannot be modeled directly using the shifted window mechanism. As shown in Figure \ref{fig: shift_wsa}, this mechanism requires a WSA and a shifted WSA to model long-term dependencies. Based on the analyses in \cite{SwinT} and \cite{SwinIR} and Equation \ref{equ: 1}, the computational complexity of model such dependencies using shifted window mechanism can be expressed as follows:
	\begin{equation}
		\begin{aligned}
			\Omega &= 2 \times \frac{HW}{kk} \times [3C^2kk + 2C(kk)^2]\\
		              &= (6C + 4N)CHW,
		\end{aligned}
	\end{equation}
where $N = k \times k$ denotes the total number of pixels in a local window. The above equation shows that the complexity of the striped window mechanism is lower than that of the shifted window mechanism, which proves its efficiency. We exclude the discussion of the overlapped window mechanism due to its excessive computational complexity. Please refer to \cite{HAT, ESRT} for more information.

\subsection{Flexible Window Training Strategy} \label{sec: 3.3}
A better training strategy is crucial for achieving better image SR results \cite{RCAN-it, RFDN, EDT, HAT, RLFN}. According to EDT \cite{EDT}, allowing models to leverage similar feature representations across the same or related tasks is essential for the success of these strategies. For instance, image SR can benefit from similar feature representation of image denoising \cite{EDT}, and $\times 4$ image SR can benefit from the similar feature representation of $\times 2$ image SR \cite{RCAN-it, RFDN}. Thus, we question whether ESWTs utilizing different striped windows (e.g., (a, b) and (c, d)) exhibit similar feature representations, and if so, whether they can benefit from each other to enhance their SR performance.

To investigate the similarity of feature representation across models, we introduced central kernel alignment (CKA) \cite{CKA}. Specifically, CKA takes two activations $\mathbf{X} \in \mathbb{R}^{m \times p_{1}}$ and $ \mathbf{Y} \in \mathbb{R}^{m \times p_{2}}$ of two layers, which contain $m$ samples and $p_1$/$p_2$ neurons, as input. The CKA of the two layers can be calculated as follows.
	\begin{equation}
		{\rm CKA} (\mathbf{K}, \mathbf{L}) = \frac{{\rm HSIC} (\mathbf{K}, \mathbf{L})}{\sqrt{ {\rm HSIC}(\mathbf{K}, \mathbf{K}) {\rm HSIC}(\mathbf{L}, \mathbf{L})}},
	\end{equation}
where $\mathbf{K} = \mathbf{X}\mathbf{X}^\top$ and $\mathbf{L} = \mathbf{Y}\mathbf{Y}^\top$ denote the Gram matrices for the two layers, and ${\rm HSIC}(\cdot)$ is the Hilbert-Schmidt independence criterion \cite{HSIC}. Given the centering matrix $\mathbf{H}=\mathbf{I}_{n}-\frac{1}{n}\mathbf{1}\mathbf{1}^\top$, $\mathbf{K}^{'}=\mathbf{H}\mathbf{K}\mathbf{H}$ and $\mathbf{L}^{'}=\mathbf{H}\mathbf{L}\mathbf{H}$ are centered Gram matrices, then we have ${\rm HSIC}(\mathbf{K}, \mathbf{L}) = {\rm vec}(\mathbf{K}^{'}) \cdot {\rm vec}(\mathbf{L}^{'})/(m-1)^{2}$. For more efficient evaluation, we use the minibatch estimator \cite{miniCKA} of CKA with a minibatch of $288$.

\begin{figure}[thbp]
    \centering
    \includegraphics[scale=.5]{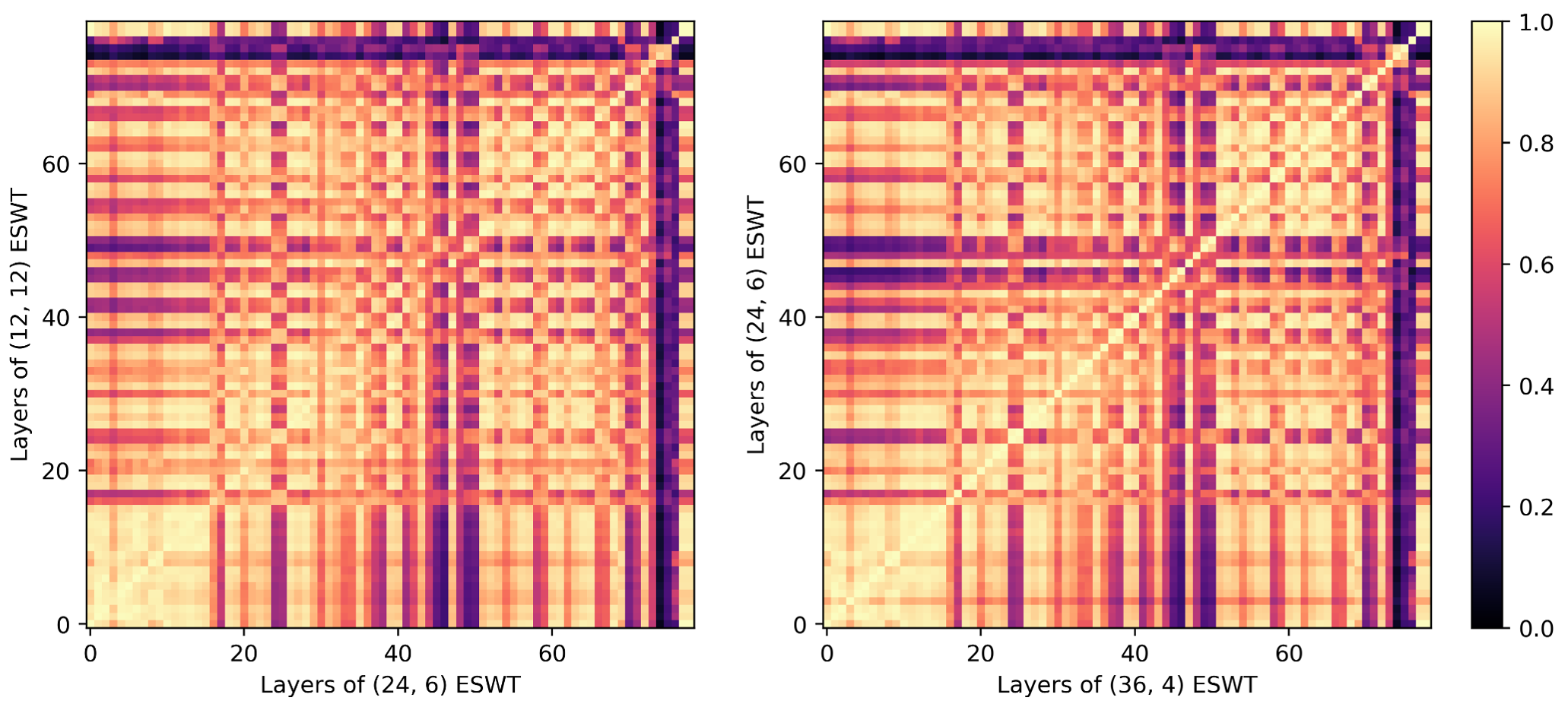}
    \caption{
		CKA similarities between all pairs of convolutional layers in ESWT using different striped windows. $X$ and $Y$ axes indexing layers from input to output. As shown on the diagonal of the heat maps, different ESWTs have \textbf{similar} feature representations.
    }
    \label{fig: cka}
\end{figure}

We train three ESWTs from scratch using the same training strategy with $(12, 12)$, $(24, 6)$, and $(36, 4)$ striped windows. The CKA similarity between all convolutional layer pairs are reported in Figure \ref{fig: cka}. We observe that ESWT using different striped windows exhibit highly similar feature representations. Based on this finding, we propose a novel flexible window training strategy that consists of several stages. The latter stage loads the pre-trained model from the previous one and calculates WSA using stretched striped windows. Experiments demonstrate that this training strategy can effectively improve the performance of ESWT without additional training cost (see Section \ref{sec: 4.3} for details).

\section{Experiments}
In this section, we first perform a series of ablation studies to validate the design choices of ESWT described above. Then, we apply the proposed ESWT to LSR to demonstrate its effectiveness. Finally, the flexible window training strategy is employed to further improve the performance of ESWT.

\subsection{Ablation studies} \label{sec: 4.1}
Following previous works \cite{LBNet, ESRT, IMDN, FDIWN, RFDN}, we conduct ablation studies using the following settings if not specified: the DF2K \cite{DIV2K} dataset selected as the training and validation set, and five benchmark datasets, namely, Set5 \cite{Set5}, Set14 \cite{Set14}, BSD100 \cite{BSD100}, Urban100 \cite{Urban100}, and Manga109 \cite{Manga109}, as the test set. The number $n$ of ETBs in ESWT is set to $3$, and the number $m$ of ETLs in each ETB is set to $6$. The striped window mechanism using an $(24,6)$ striped window. We train models by minimizing the L1 loss using the Adam optimizer \cite{Adam} with $\beta_{1}=0.9$ and $\beta_{2}=0.999$ for a total of $100K$ iterations. The initial learning rate is set to $5 \times 10^{-4}$ and it is decreased to $5 \times 10^{-6}$ using the cosine annealing scheduler \cite{Cosine}. The training batch size is $64$ and the training patch size is $72 \times 72$. Random rotation and random horizontal flip are used as data enhancement.

For evaluation of the quality of the SR images, the PSNR and SSIM calculated on the Y channel of the YCbCr space are used. For a more comprehensive evaluation, four metrics are used to measure model complexity, ``\#Params'' the total number of learnable parameter, ``\#Memory'' the maximum GPU memory consumption, ``\#Latency'' the average inference time per image on a dataset, ``\#FLOPs'' the abbreviation for floating point operations. To ensure a fair comparison, we report all ``\#FLOPs'' using a $3 \times 256 \times 256$ image for $\times 4$ image SR.

\begin{table}[htbp]
  \centering
  \resizebox{0.85\linewidth}{!}{
    \begin{tabular}{l|c|cc|cc}

    \hline
    \specialrule{0em}{2pt}{0pt}

       Initial & Normalization & \multicolumn{2}{c|}{Urban100}        & \multicolumn{2}{c}{Manga109}   \\
       Learning Rate & Layer & PSNR & SSIM & PSNR & SSIM \\

    \specialrule{0em}{1pt}{0pt}
    \hline
    \specialrule{0em}{1pt}{0pt}
    \hline
    \specialrule{0em}{2pt}{0pt}

    \multirow{6}{*}{1e-4} &Identity & 23.29  & 0.6985  & 27.92  & 0.8434  \\
    	   &Identity$^\dagger$ & 25.12  & 0.7443  & 28.63  & 0.8775  \\
           &LN& 25.11  & 0.7526  & 28.59  & 0.8800  \\
           &BN& 25.22  & 0.7910  & 28.73  & 0.8831  \\
           &Single BN& 25.27  & 0.7917  & 28.77  & 0.8843  \\
               \rowcolor{Graylight}
           &Embedded BN& 25.49  & 0.7923  & 29.11  & 0.8901  \\

    \specialrule{0em}{1pt}{0pt}
    \hline
    \specialrule{0em}{2pt}{0pt}

    \multirow{5}{*}{2e-4}  &Identity$^\ddagger$& - & - & - & - \\
           &LN& 26.23  & 0.7905  & 30.60  & 0.9099  \\
           &BN& 26.31  & 0.7933  & 30.69  & 0.9113  \\
           &Single BN& 26.36  & 0.7951  & 30.75  & 0.9120  \\
               \rowcolor{Graylight}
           &Embedded BN& 26.45  & 0.7974  & 30.87  & 0.9126  \\

    \specialrule{0em}{1pt}{0pt}
    \hline
    \specialrule{0em}{2pt}{0pt}

    \multirow{5}{*}{5e-4}  &Identity$^\ddagger$& - & - & - & - \\
           &LN& 26.41  & 0.7955  & 30.83  & 0.9125  \\
           &BN& 26.45  & 0.7970  & 30.87  & 0.9133  \\
           &Single BN& 26.49  & 0.7991  & 30.89  & 0.9133  \\
               \rowcolor{Graylight}
           &Embedded BN& 26.54  & 0.8013  & 30.89  & 0.9137  \\

    \specialrule{0em}{1pt}{0pt}
    \hline
    \specialrule{0em}{1pt}{0pt}
    \hline
    \specialrule{0em}{2pt}{0pt}

    \end{tabular}
  }
  \caption{
	Ablation study about the role of normalization layers in shallow transformers. ``Identity'' means no normalization layer is used. ``Single BN'' means the BN layer is used before SA. ``Embedded BN'' means the BN layer is embedded into SA. ``$\dagger$'' means the use of pre-training and fine-tuning strategies in \cite{RCAN-it} and training for $\times 5$ total iterations. ``$\ddagger$'' means the training is unstable.
  }
  \label{tab: as_tl}
\end{table}
\begin{table}[htbp]
    \centering
    \resizebox{0.85\linewidth}{!}{
        \begin{tabular}{l|cccc}

        \hline
        \specialrule{0em}{2pt}{0pt}

        Normalization      & Params          & FLOPs          & Memory          & Latency \\
                       Layer                 & [K]              & [G]            & [M]             & [ms] \\

        \specialrule{0em}{1pt}{0pt}
        \hline
        \specialrule{0em}{1pt}{0pt}
        \hline
        \specialrule{0em}{2pt}{0pt}

    Identity  & 585 & 36.84 & 116.71 & 760.98 \\
    LN    & 589 & 37.91 & 142.84 & 820.54 \\
    BN    & 589 & 38.19 & 122.36 & 820.59 \\
    Single BN & 587 & 38.05 & 119.06 & 820.55 \\
    \rowcolor{Graylight}
    Embedded BN   & 589 & 38.19 & 121.82 & 760.08 \\

        \specialrule{0em}{1pt}{0pt}
        \hline
        \specialrule{0em}{1pt}{0pt}
        \hline
        \specialrule{0em}{2pt}{0pt}

        \end{tabular}
    }
	\caption{
		Model complexity of $\times 4$ image SR on the DIV2K \cite{DIV2K} validation dataset using different normalization layers.
	}
	\label{tab: as_etl}
\end{table}
\begin{table}[htbp]
    \centering
    \resizebox{0.85\linewidth}{!}{
        \begin{tabular}{l|c|cc|cc|cc}

        \hline
        \specialrule{0em}{2pt}{0pt}

        \multirow{2}{*}{Mechanism}               & \multirow{2}{*}{Window Size}             & \multicolumn{2}{c|}{BSD100} & \multicolumn{2}{c|}{Urban100} & \multicolumn{2}{c}{Manga109} \\
                                                 &                                          & PSNR                        & SSIM                          & PSNR                            & SSIM            & PSNR           & SSIM  \\

        \specialrule{0em}{1pt}{0pt}
        \hline
        \specialrule{0em}{1pt}{0pt}
        \hline
        \specialrule{0em}{2pt}{0pt}

        w/o Striped Window                        &(12, 12)                                 & 27.69                       & 0.7404                        & 26.48                           & 0.7977          & 30.88          & 0.9129  \\

        \specialrule{0em}{1pt}{0pt}
        \hline
        \specialrule{0em}{2pt}{0pt}

        \multirow{3}{*}{w/ Striped Window}           &(16, 9)                                  & 27.69                       & 0.7409                        & 26.49                           & 0.7987          & 30.91          & 0.9130  \\
                                                  &(18, 8)                                  & 27.69                       & 0.7407                        & 26.51                           & 0.7990          & 30.90          & 0.9134  \\
                                                      \rowcolor{Graylight}
                                                  &(24, 6)                                  & \textbf{27.70}              & \textbf{0.7410}               & \textbf{26.56}                  & \textbf{0.8006} & \textbf{30.94} & \textbf{0.9136} \\

        \specialrule{0em}{1pt}{0pt}
        \hline
        \specialrule{0em}{1pt}{0pt}
        \hline
        \specialrule{0em}{2pt}{0pt}

        \end{tabular}
    }
	\caption{
		Ablation study about the effectiveness of striped window mechanism. The study was performed on the BSD100 \cite{BSD100}, Urban100 \cite{Urban100}, and Manga109 \cite{Manga109} datasets at $\times 4$ image SR. The BEST results are \textbf{highlighted}.
	}
	\label{tab: as_2}
\end{table}

\paragraph{From LN to Embedded BN}
We first examine the impact of the normalization layer in shallow transformers. The results of the ablation study are shown in Table \ref{tab: as_tl}. Our key findings can be summarized as follows: Although removing the normalization layer in ESWT at low learning rates can decrease model complexity (Table \ref{tab: as_etl}) without compromising performance, this strategy incurs significant training costs and restricts the transformer from fully exploiting the performance gains achievable with higher learning rates \cite{lr1, lr2, lr3}. In addition, although EDSR \cite{EDSR} has claimed that BN is unsuitable for SR models with high model capacity due to its tendency to remove differences between low-level visual features (such as edges and textures) across samples in the mini-batch, our experiments indicate that incorporating BN can improve the generalization capability of shallow transformers with limited capacity more effectively than LN. This finding is particularly noteworthy given that our model accounts for only 1.4\% of the parameter number of EDSR. We also confirm that excessive normalization layers may penalize the performance of models with limited capacity \cite{StyleGAN2, ESRGCNN, b-v1, b-v2}. Finally, inspired by \cite{ELAN}, we embed the BN layer into SA and observe that it can further improve the model performance and has a low model complexity. To sum up, we suggest the removal of the normalization layer in TL and incorporation of BSA to improve the generalization capability and performance of shallow transformers, while maintaining a balance between model complexity, performance, and training cost.

Given that this research is a basic ablation study, we retain a number of fixed factors that can be investigated further, including the normalization layer location and model parameters. We will consider them for future exploration.

\begin{table*}[thbp]
  \scriptsize
  \centering
  \resizebox{0.85\linewidth}{!}{
    \begin{tabular}{l|c|c|cc|cc|cc|cc|cc}

    \hline
    \specialrule{0em}{1pt}{0pt}

    \multirow{2}{*}{Method}     & \multirow{2}{*}{Scale}      & \multirow{2}{*}{Params} & \multicolumn{2}{c|}{Set5} & \multicolumn{2}{c|}{Set14} & \multicolumn{2}{c|}{BSD100} & \multicolumn{2}{c|}{Urban100} & \multicolumn{2}{c}{Manga109} \\
                                &                             &                         & PSNR                      & SSIM                       & PSNR                        & SSIM                          & PSNR                            & SSIM            & PSNR           & SSIM            & PSNR           & SSIM \\

    \specialrule{0em}{1pt}{0pt}
    \hline
    \specialrule{0em}{1pt}{0pt}
    \hline
    \specialrule{0em}{2pt}{0pt}

    ESRT\cite{ESRT}             & \multirow{6}{*}{$\times3$}  & 770                     & 34.42                     & 0.9268                     & 30.43                       & 0.8433                        & 29.15                           & 0.8063          & 28.46          & 0.8574          & 33.95          & 0.9455  \\
    LBNet\cite{LBNet}           &                             & 736                     & 34.47                     & 0.9277                     & 30.38                       & 0.8417                        & 29.13                           & 0.8061          & 28.42          & 0.8559          & 33.82          & 0.9460  \\
    SwinIR\cite{SwinIR}         &                             & 886                     & 34.62                     & 0.9289                     & 30.54                       & 0.8463                        & 29.20                           & 0.8082          & 28.66          & 0.8624          & 33.98          & 0.9478  \\
    ELAN-light\cite{ELAN}       &                             & \underline{590}                     & 34.61                     & 0.9288                     & 30.55                       & 0.8463                        & 29.21                           & 0.8081          & 28.69          & 0.8624          & 34.00          & 0.9478  \\
    \rowcolor{Graylight}
    ESWT(ours)                  &                             & \textbf{578}                     & \underline{34.63}            & \underline{0.9290}            & \underline{30.55}              & \underline{0.8464}               & \underline{29.23}                  & \underline{0.8088} & \underline{28.70} & \underline{0.8628} & \underline{34.05} & \underline{0.9479} \\
    \rowcolor{Graylight}
    ESWT$^\dagger$(ours)                  &                             & \textbf{578}                     & \textbf{34.65}            & \textbf{0.9301}            & \textbf{30.56}              & \textbf{0.8467}               & \textbf{29.24}                  & \textbf{0.8097} & \textbf{28.71} & \textbf{0.8633} & \textbf{34.07} & \textbf{0.9480} \\

    \specialrule{0em}{1pt}{0pt}
    \hline
    \specialrule{0em}{2pt}{0pt}

    ESRT\cite{ESRT}             & \multirow{6}{*}{$\times4$}  & 751                     & 32.19                     & 0.8947                     & 28.69                       & 0.7833                        & 27.69                           & 0.7379          & 26.39          & 0.7962          & 30.75          & 0.9100  \\
    LBNet\cite{LBNet}           &                             & 742                     & 32.29                     & 0.8960                     & 28.68                       & 0.7832                        & 27.62                           & 0.7382          & 26.27          & 0.7906          & 30.76          & 0.9111  \\
    SwinIR\cite{SwinIR}         &                             & 897                     & 32.44                     & 0.8976                     & 28.77                       & 0.7858                        & 27.69                           & 0.7406          & 26.47          & 0.7980          & 30.92          & \textbf{0.9151}  \\
    ELAN-light\cite{ELAN}       &                             & \underline{601}                     & 32.43                     & 0.8975                     & 28.78                       & 0.7858                        & 27.69                           & 0.7406          & 26.54          & 0.7982          & 30.92          & \underline{0.9150}  \\
    \rowcolor{Graylight}
    ESWT(ours)                  &                             & \textbf{589}                     & \underline{32.46}            & \underline{0.8979}            & \underline{28.80}              & \underline{0.7866}               & \underline{27.70}                  & \underline{0.7410} & \underline{26.56} & \underline{0.8006} & \underline{30.94} & 0.9136  \\
    \rowcolor{Graylight}
    ESWT$^\dagger$(ours)                  &                             & \textbf{589}                     & \textbf{32.48}            & \textbf{0.8982}            & \textbf{28.82}              & \textbf{0.7869}               & \textbf{27.71}                  & \textbf{0.7415} & \textbf{26.57} & \textbf{0.8011} & \textbf{30.95} & 0.9143  \\

    \specialrule{0em}{1pt}{0pt}
    \hline
    \specialrule{0em}{1pt}{0pt}
    \hline
    \specialrule{0em}{2pt}{0pt}

    \end{tabular}
  }
  \caption{Quantitative comparison of $\times 3$ and $\times 4$ image SR. The BEST and second BEST results are \textbf{highlighted} and \underline{underlined}. ``$\dagger$'' means the use of flexible window training strategy.
  }
  \label{tab: comparison}
\end{table*}

\begin{figure*}[thbp]
    \centering
    \includegraphics[scale=.3]{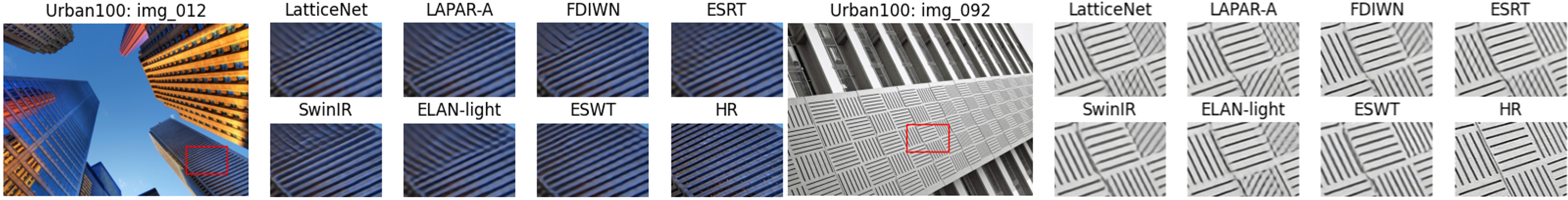}
    \caption{
    	Qualitative comparison of $\times 4$ image SR on Urban100 \cite{Urban100} dataset.
    }
    \label{fig: comparison}
\end{figure*}
\begin{table}[htbp]
    \centering
    \resizebox{0.85\linewidth}{!}{
        \begin{tabular}{l|cc|cccc}

        \hline
        \specialrule{0em}{2pt}{0pt}

        \multirow{2}{*}{Method}         & \multicolumn{2}{c|}{BSD100}   & Params           & FLOPs         & Memory          & Latency  \\
                                        & PSNR                          & SSIM             & [K]           & [G]             & [M]             & [ms] \\

        \specialrule{0em}{1pt}{0pt}
        \hline
        \specialrule{0em}{1pt}{0pt}
        \hline
        \specialrule{0em}{2pt}{0pt}

        ESRT                            & 27.69                         & 0.7379           & 752           & 75.83           & 584.80          & 20.16  \\
        LBNet                           & 27.62                         & 0.7382           & 742           & 173.26          & 144.04          & 25.41  \\
        SwinIR                          & 27.69                         & 0.7406           & 897           & 218.78          & 71.90           & 34.25  \\
        ELAN-light                      & 27.69                         & 0.7406           & 601           & 39.09           & 45.14           & 15.22  \\
\rowcolor{Graylight}
        ESWT(ours)                      & \textbf{27.70}                & \textbf{0.7410}  & \textbf{589}  & \textbf{38.20}  & \textbf{37.60}  & \textbf{12.08}  \\

        \specialrule{0em}{1pt}{0pt}
        \hline
        \specialrule{0em}{1pt}{0pt}
        \hline
        \specialrule{0em}{2pt}{0pt}

        \end{tabular}
    }
    \caption{
    	Quantitative trade-off comparison between model performance and complexity of $\times 4$ image SR on BSD100 \cite{BSD100} dataset. The BEST results are \textbf{highlighted}.
    }
    \label{tab: trade_off}
\end{table}

\begin{table}[htbp]
    \centering
    \resizebox{0.85\linewidth}{!}{
        \begin{tabular}{l|c|c|c}

        \hline
        \specialrule{0em}{2pt}{0pt}

        Stage      & Window Size          & Learning Rate          & iterations   \\

        \specialrule{0em}{1pt}{0pt}
        \hline
        \specialrule{0em}{1pt}{0pt}
        \hline
        \specialrule{0em}{2pt}{0pt}

    One  & (12, 12) & $5 \times 10^{-4} \rightarrow 5 \times 10^{-6}$ & 50K  \\
    Two    & (18, 8) & $1 \times 10^{-4} \rightarrow 1 \times 10^{-6}$ & 25K   \\
    Three    & (24, 6) & $1 \times 10^{-4} \rightarrow 1 \times 10^{-6}$ & 25K  \\

        \specialrule{0em}{1pt}{0pt}
        \hline
        \specialrule{0em}{1pt}{0pt}
        \hline
        \specialrule{0em}{2pt}{0pt}

        \end{tabular}
    }
	\caption{
	    A simple three-stage flexible window training strategy. The latter stage loads the pre-training weights from the previous stage to benefit from the learned feature representations.
	}
	\label{tab: ts}
\end{table}

\begin{figure}[htbp]
    \centering
    \includegraphics[scale=0.22]{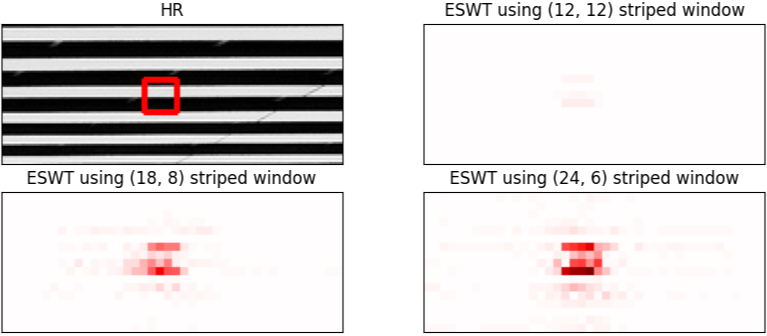}
    \caption{
        LAM results of ESWT using different striped windows in $\times 4$ image SR. When reconstructing the patches marked with red boxes, a dark color indicates a larger degree of contribution.
        }
    \label{fig: lam}
\end{figure}

\begin{figure}[htbp]
    \centering
    \includegraphics[scale=0.385]{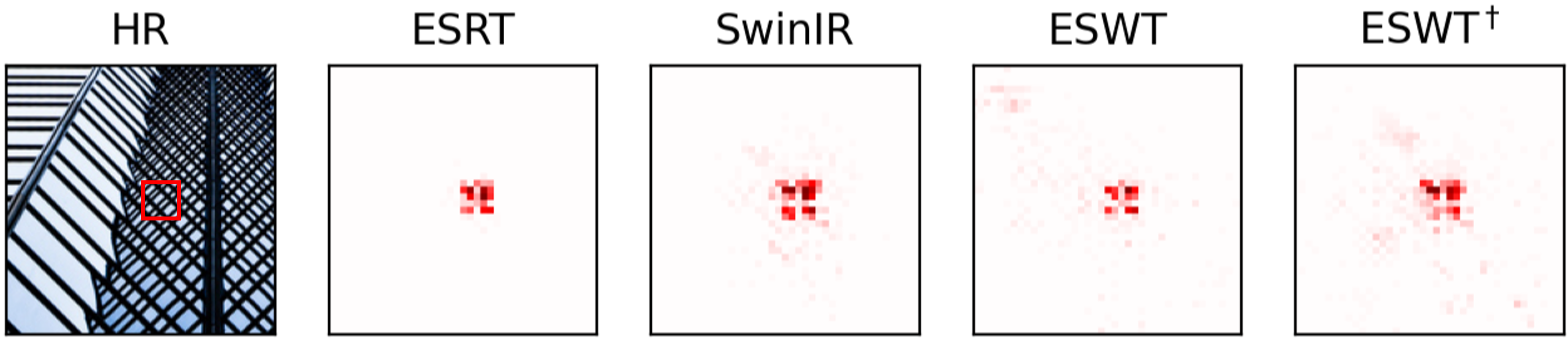}
    \caption{
        LAM results of different SR models in $\times 4$ image SR. ESRT uses the overlapped window mechanism and SwinIR uses the shifted window mechanism. ESWT using striped window mechanism and/or flexible window training strategy can explore a \textbf{larger} range of contextual information. ``$\dagger$'' means the use of flexible window training strategy.
        }
    \label{fig: lam2}
\end{figure}

\paragraph{Effectiveness of Striped Window Mechanism}
In Section \ref{sec: 3.2}, we propose a striped window mechanism to model long-term dependencies effectively. Table \ref{tab: as_2} reports the performance of ESWT on $\times 4$ image SR using different mechanisms and striped windows. For fair comparison, all local windows contain $144$ pixels. The results of the ablation study indicate that the striped window mechanism can improve the performance of the ESWT. Moreover, the stretching of striped window can further enhance the model performance. To better understand the proposed striped window mechanism, we introduce local attribution map (LAM) for overall model analysis. Suppose that $F: \mathbb{R}^{h\times w}\mapsto\mathbb{R}^{sh\times sw}$ is an image SR model with $\times s$ upscaling factor, LAM employs the path integral gradient for its attribution analysis \cite{LAM}. The LAM results for the ESWT using different striped windows are shown in Figure \ref{fig: lam} and the LAM results for the model using different mechanisms are shown in Figure \ref{fig: lam2}. It can be observed that by incorporating $(24, 6)$ striped windows, the ESWT can make more efficient utilization of contextual information. To further investigate the effect of the proposed mechanism on the integration of contextual information in the transformer, we analyze the mean attention distance (MAD) at different SA layers, which is analogous to receptive field in CNNs. Figure \ref{fig: mad} shows that the proposed mechanism improves the MAD of the shallow SA layers of the ESWT and thus enhances its perception of contextual information. These analysis results confirm that the striped window mechanism contributes to better exploration of contextual information and thus improves model performance. As the official pre-training weights had not been released, we retrained the ESRT used to calculate LAM using the training settings reported in its paper \cite{ESRT}.

\subsection{Application}
We now apply ESWT to LSR following the training settings of ablation study and compare it with state-of-the-art (SOTA) methods in terms of the following aspects: qualitative and quantitative results of image SR, and the trade-off between model complexity and performance.

\paragraph{Quantitative and Qualitative Comparisons}
The quantitative comparison results are shown in Table \ref{tab: comparison}. The proposed ESWT performs better in almost all cases, surpassing ELAN \cite{ELAN} 0.0024 SSIM on the Urban100 \cite{Urban100} dataset of $\times 4$ image SR. The qualitative comparison results are illustrated in Figure \ref{fig: comparison}. Compared with other methods, the SR result reconstructed by the proposed ESWT contains rich structural details, has sharper edges, and looks more natural.

\paragraph{Model Performance and Complexity Trade-off}
In the evaluation of LSR methods, the trade-off between model performance and complexity is also a key factor to consider. The qualitative trade-off comparison between ESWT and SOTA methods is shown in Table \ref{tab: trade_off}. The proposed ESWT reduces the complexity of the model and further enhances the quality of SR images, achieving a better trade-off compared with SOTA methods.

\subsection{Training with Flexible Window} \label{sec: 4.3}

\begin{figure}[htbp]
    \centering
    \includegraphics[scale=0.39]{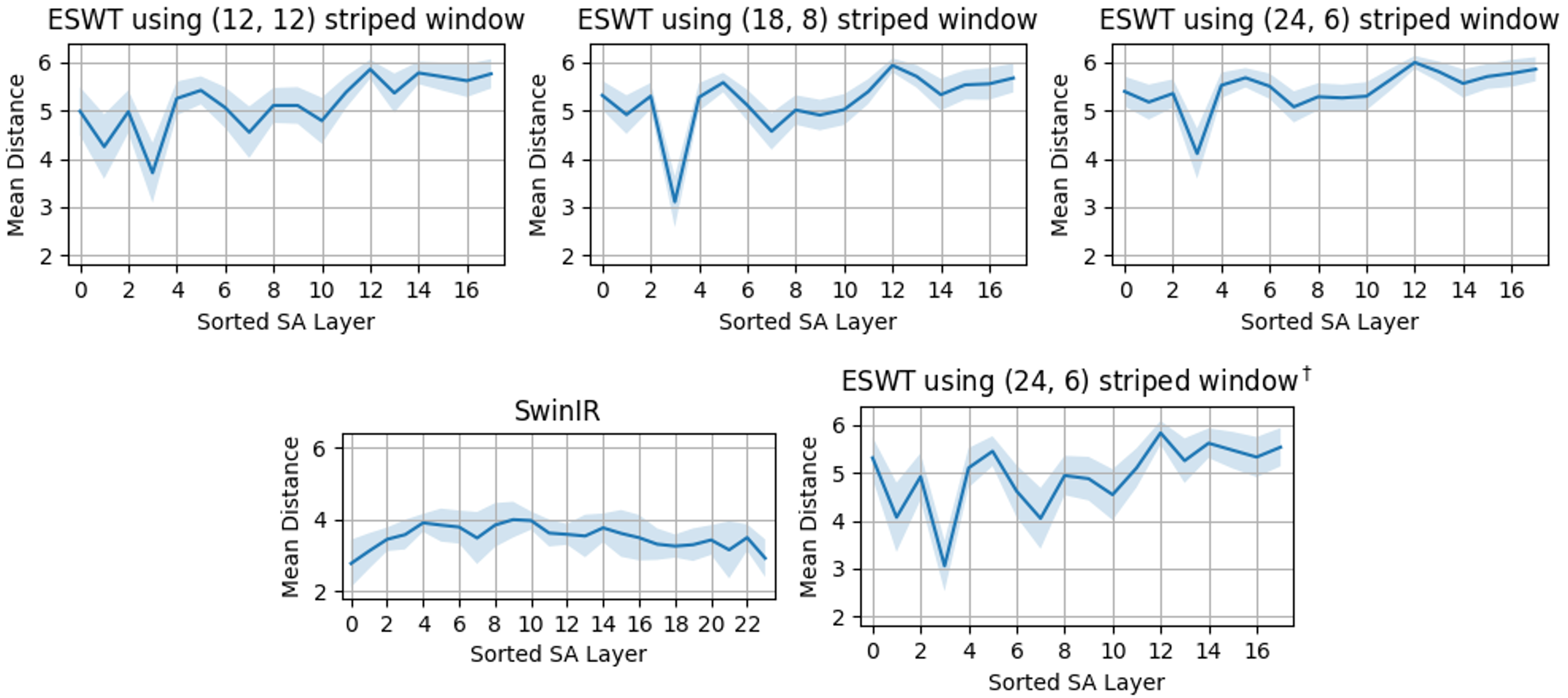}
    \caption{
    	MAAs (light blue areas) of SwinIR and ESWT using different striped windows in $\times 4$ image SR. MAAs are determined by the mean and standard deviation of $100$ MADs (blue lines) obtained on the DIV2K \cite{DIV2K} validation set. MADs are obtained by averaging the distances between query pixels and all other pixels, weighted by the attention weights \cite{1616, miniCKA}. Due to hardware limitation, we do not analyze the MAA of ESRT. ``$\dagger$'' means the use of flexible window training strategy.
    }
    \label{fig: mad}
\end{figure}
\begin{figure}[thbp]
    \centering
    \includegraphics[scale=.32]{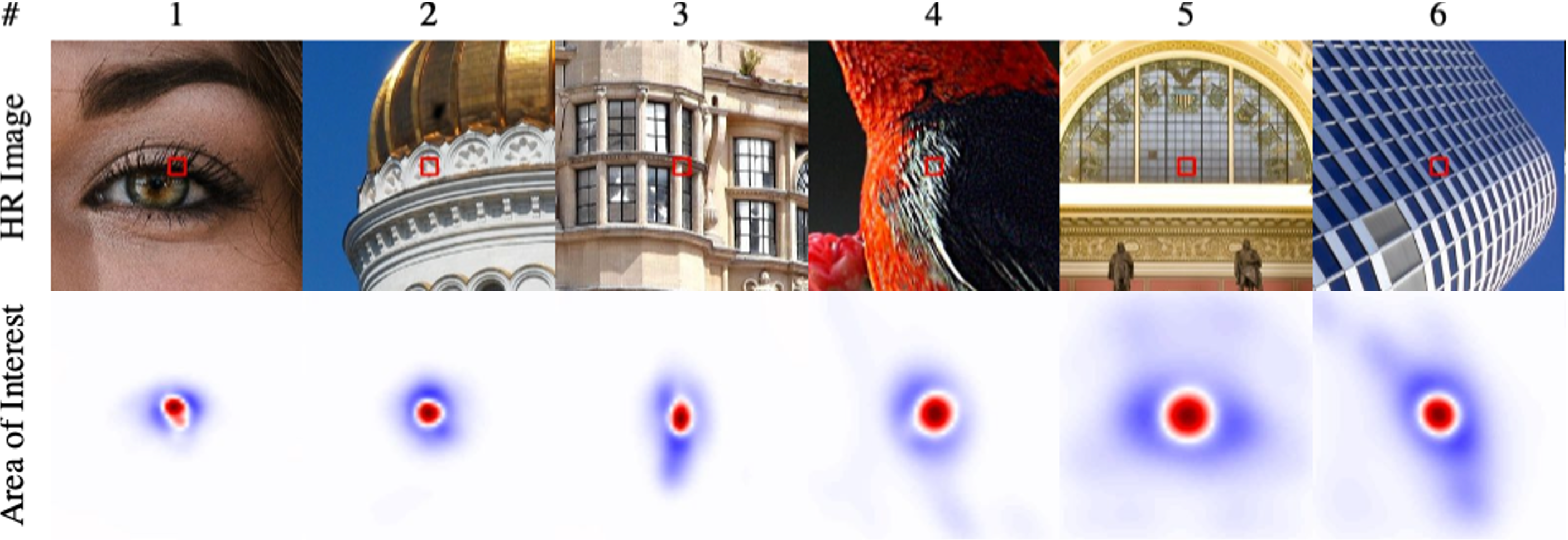}
    \caption{
        Areas of interest across different SR models. \textcolor{red}{Red} and \textcolor{blue}{blue} regions both contribute to the SR images, but the \textcolor{blue}{blue} regions are more challenging for SR models to utilize. Samples are selected from \cite{LAM}.
    }
    \label{fig: in}
\end{figure}

High-quality SR images are obtained through joint optimization of the model, data, and training strategy. In Section \ref{sec: 3.3}, we analyze the effectiveness of existing training strategies and propose a new flexible window training strategy. We now apply the proposed training strategy on ESWT to verify its advancement. Referring to RLFN \cite{RLFN}, we design a simple three-stage training strategy (Table \ref{tab: ts}). In the first stage, a high learning rate is used to improve the generalization capability of the model. In subsequent stages, the striped window is stretched to expand the receptive field of the model, and a lower learning rate is used for the model to fine-tune its feature representation for better SR performance. We maintain the same training settings in the ablation study except those specified in the Table \ref{tab: ts}.

Qualitative experimental results are presented in Table \ref{tab: comparison}. The proposed flexible window training strategy improves the performance of ESWT considerably without any additional training costs. The following reasons may contribute to its effectiveness: Given that ESWTs with different striped windows have highly similar feature representations (Figure \ref{fig: cka}), by loading the pre-training weight of the previous stage, the latter stage can benefit from shared feature representations. Moreover, we analyzed the mean attention area (MAA) of the SA layer in different models, which indicates the most important areas attended by SA \cite{EDT, miniCKA, 1616} (Figure \ref{fig: mad}). The proposed strategy broadens the MAA of SA layers, indicating that it enables ESWT to incorporate more contextual information to build high-quality SR images. Furthermore, Gu et al. \cite{LAM} observed a difficult-to-learn contextual information boundary for SR models (Figure \ref{fig: in}). The proposed flexible windowing strategy enables ESWT to vary its receptive fields during different stages of training, which allows it to initially capture contextual information at local scales and subsequently leverage it to better model dependencies at larger scales. Compared with increasing the depth of the model \cite{SwinIR} and using overlapped WSA \cite{ESRT} to enlarge the receptive field, we consider that the proposed training strategy is an effective method for crossing this boundary.

\section{Conclusions}
In this paper, we propose an efficient striped window transformer (ESWT) for lightweight super-resolution (LSR). Specifically, by removing the normalization layer and embedding batch normalization (BN) in self-attention (SA), we propose an efficient backbone for shallow transformers, which gives the proposed ESWT a concise structure. In addition, we propose a striped window mechanism that exhibits superior long-term dependency modeling capability and lower complexity than the shifted and overlapped window mechanisms. Furthermore, we propose a flexible window training strategy. By dynamically adjusting the receptive field at different stages of training and sharing the learned feature representation, the proposed strategy assists the ESWT to better utilize contextual information without any additional cost. Extensive experiments reveal that the proposed method outperforms SOTA methods with fewer parameters, faster inference, smaller FLOPs, and less memory consumption, achieving a better trade-off between model performance and complexity.

{\small
\bibliographystyle{ieee_fullname}
\bibliography{iccv.bib}
}

\end{document}